\documentclass{IOS-Book-Article}

\usepackage{mathptmx}
\usepackage{soul}\setuldepth{article}

\usepackage{graphicx}
\usepackage{multicol}
\usepackage{booktabs}
\usepackage{lipsum}
\usepackage[numbers, sort&compress]{natbib}

\def\hb{\hbox to 11.5 cm{}}

\begin{document}

\pagestyle{headings}
\def\thepage{}
\begin{frontmatter} 

\title{Secondary Use of Clinical Problem List Entries for Neural Network-Based Disease Code Assignment}

\markboth{}{Jan 2023\hb}

\author[A]{\fnms{Markus} \snm{Kreuzthaler}
\thanks{Corresponding Author: Markus Kreuzthaler, markus.kreuzthaler@medunigraz.at}},
\author[A]{\fnms{Bastian} \snm{Pfeifer}},
\author[B]{\fnms{Diether} \snm{Kramer}}
and
\author[A]{\fnms{Stefan} \snm{Schulz}}

\runningauthor{M. Kreuzthaler et al.}
\address[A]{Institute for Medical Informatics, Statistics and Documentation, \\Medical University of Graz, Austria}
\address[B]{Department of Information and Process Management, \\Steiermärkische Krankenanstaltengesellschaft m.b.H. (KAGes), Graz, Austria}

\begin{abstract}
Clinical information systems have become large repositories for semi-structured and partly annotated electronic health record data, which have reached a critical mass that makes them interesting for supervised data-driven neural network approaches. We explored automated coding of 50 character long clinical problem list entries using the International Classification of Diseases (ICD-10) and evaluated three different types of network architectures on the top 100 ICD-10 three-digit codes. A fastText baseline reached a macro-averaged F1-score of 0.83, followed by a character-level LSTM with a macro-averaged F1-score of 0.84. The top performing approach used a downstreamed RoBERTa model with a custom language model, yielding a macro-averaged F1-score of 0.88. A neural network activation analysis together with an investigation of the false positives and false negatives unveiled inconsistent manual coding as a main limiting factor.
\end{abstract}

\end{frontmatter}
\markboth{April 2022\hb}{April 2022\hb}

\section{Introduction and Motivation}
\label{subsec:introduction}
The clinical information system (CIS) of a large, multicentre public hospital provider in Austria stores short clinical problem descriptions in German (maximum 50 characters) together with which manually assigned codes from the International Classification of Diseases (ICD-10). This huge table fulfils three purposes: (i) collection of content that automatically fills the ``Diagnoses'' section when narrative discharge summaries are created, (ii) display of a problem-list like scrollable textbox in the CIS frontend, and (iii) provision of ICD codes for administrative purposes. Due to the technical limitation of 50 characters, the often-lengthy ICD-10 descriptions are usually overwritten by the users, who use overly compact expressions, characterized by ellipsis, context-dependent abbreviations and acronyms, non-standardised numeric values, spelling variants and errors. These text snippets exemplify typical idiosyncrasies of clinical language~\citep{Neveol2018}.

Our work is centred on this resource. We wanted to investigate to what extent clinical real-world data is suited as training material for different types of multi-class classification approaches for automatic assignment of codes from the classification system ICD-10, and where it reaches its limits. We applied three different types of neural network (NN) architectures: a shallow NN, a recurrent NN and a transformer-based architecture for our experimental secondary use-case scenario of clinical routine documentation.

\section{Methods and Data}
\label{subsec:methodsAndData}
\subsection{Data Set}
\label{subsec:dataSet}
We used $\sim$1.9 million unique de-identified problem list entries and ignored all entries without ICD codes. A 90/10 split was carried out for training and test set preparation. We trained one model using the top 100 occurring three-digit ICD-10 codes, which cover about 50\% of all code instances in the data set. Therefore, we scaled up all training samples of this highly imbalanced data set to the most frequent code, resulting in $\sim$6 million training samples (100 times $\sim$60k observations). The test set based distribution of the 100 codes under investigation remained unchanged ($\sim$93k observations).

\subsection{Neural Network Architectures}
\label{subsec:neuralNetwork}

\begin{description}
    \item[Shallow.] As a baseline, we used \textit{fastText}~\citep{Bojanowski2016a}, exploiting pre-trained skip-gram embeddings on a subword and word type level from the training set. We used a simple rule-based tokenizer\footnote{[\textasciicircum \textbackslash p{IsAlphabetic}\textbackslash p{IsDigit}]} and normalized all resulting tokens to lower case.
    
    \item[Recurrent.] Character inputs are modelled as 122 dimensional one-hot encoded vectors from the training set, together with an out-of-character-dictionary feature dimension in accordance to~\citet{Zhang2015} for the input to the chosen LSTM~\cite{Hochreiter1997} network. \textit{Deeplearning4j} was used as implementation library.
    
    \item[Transformer.] For transformer-based architectures, we decided to apply RoBERTa~\citep{Liu2019}. We build our own language model, in order to support the downstream task with a first understanding of the language under scrutiny. To this end, we used \textit{ktrain}, a lightweight wrapper library, for \textit{Hugging Face}.
\end{description}

\subsection{Neural Network Interpretation}
\label{subsec:neuralNetworkInterpretation}
In health care scenarios, regulatory frameworks require that decision support systems are able to explain the path that led to a certain suggestion. This is optimally met by decision trees, but poses complex challenges for NNs. Established approaches are LIME (Local Interpretable Model-Agnostic Explanations), SHAP (SHapley Additive exPlanations) or the notion of saliency as being the norm of the gradient of the loss function to a given input, an approach successfully applied in clinical NLP domains~\citep{gehrmann2018comparing} for explainable ML systems. In this work, we were particularly interested in character-wise feedback for the overall classification result via the inspection of certain class probabilities at certain positions of the LSTM sequence model. This allows the identification of activation levels at this granular input representation scheme, because single characters have significant impact on the correct interpretation of narrative content.

\subsection{Evaluation Metrics}
\label{subsec:evaluationMetrics}
We evaluated the performance of the trained model on the test data set using precision, recall, and F1-score per ICD-10 code, as well as macro-evaluation statistics. The following definitions were used: true positives (\textit{TPs}) – the number of correctly assigned codes; true negatives (\textit{TNs}) – the number of correctly unassigned codes; false positives (\textit{FPs}) – the number of incorrectly assigned codes; false negatives (\textit{FNs}) – the number of incorrectly unassigned codes. Exploiting this definition, precision \textit{P = TP / (TP + FP)}, recall \textit{R = TP / (TP + FN)} and F1-score = \textit{2 $\cdot$ P $\cdot$ R / (P + R)}.

\section{Results and Discussion}
\label{subsec:resultsAndDiscussion}

\begin{table*}[!htbp]
\centering
\scriptsize 
\begin{tabular}{ll}
    \begin{tabular}{c c c c}
    \toprule
    ICD-10 & Precision & Recall & F1-score \\
    \midrule
P07 & 0,99 & 1,00 & 0,99 \\
F17 & 0,99 & 0,97 & 0,98 \\
A46 & 0,96 & 0,97 & 0,97 \\
E78 & 0,96 & 0,97 & 0,97 \\
G47 & 0,97 & 0,98 & 0,97 \\
I35 & 0,97 & 0,97 & 0,97 \\
I71 & 0,95 & 0,98 & 0,97 \\
D64 & 0,95 & 0,97 & 0,96 \\
G40 & 0,96 & 0,96 & 0,96 \\
J44 & 0,96 & 0,96 & 0,96 \\
    \bottomrule
    \end{tabular}

    \begin{tabular}{c c c c}
    \toprule
    ICD-10 & Precision & Recall & F1-score \\
    \midrule
D48 & 0,79 & 0,71 & 0,74 \\
D37 & 0,74 & 0,73 & 0,73 \\
I47 & 0,67 & 0,77 & 0,72 \\
T14 & 0,83 & 0,64 & 0,72 \\
I64 & 0,56 & 0,72 & 0,63 \\
E10 & 0,53 & 0,73 & 0,61 \\
N19 & 0,43 & 0,73 & 0,54 \\
E14 & 0,45 & 0,51 & 0,47 \\
C80 & 0,40 & 0,55 & 0,46 \\
Z03 & 0,38 & 0,44 & 0,41 \\
    \bottomrule
    \end{tabular}
    
    \end{tabular}
    \caption{Selection of test set based evaluation results using RoBERTa.}
    \label{tab:resultstestRoBERTa}
\end{table*}
\normalsize

Evaluation of the test set with 92,832 problem list entries yielded an overall macro-averaged F1-score of 0.83 for the fastText baseline, 0.84 for the LSTM, and 0.88 for the RoBERTa approach. Table~\ref{tab:resultstestRoBERTa} shows the evaluation on an ICD-10 three-digit level. The left side of the table shows the top 10 best performing codes, the right side shows codes with a macro-averaged F1-score less than 0.75. We got the highest F1-score value for the ICD-10 code P07 (``disorders related to short gestation and low birth weight, not elsewhere classified'') and the lowest one for Z03 (``medical observation and evaluation for suspected diseases and conditions'') for all three different NN architectures. Interestingly, LSTM, which solely relies on character-level inputs, showed about the same performance as the fastText baseline. This provides strong evidence that the LSTM network is highly sensitive to even minor variations of character sequences that are relevant for the correct ICD code assignment. We decided to investigate this further by performing a network activation analysis of the LSTM.

\subsection{Explainable AI}
\label{subsec:explainableAI}

\begin{figure} [htbp]
  \begin{center}
    \includegraphics[width=0.65\textwidth]{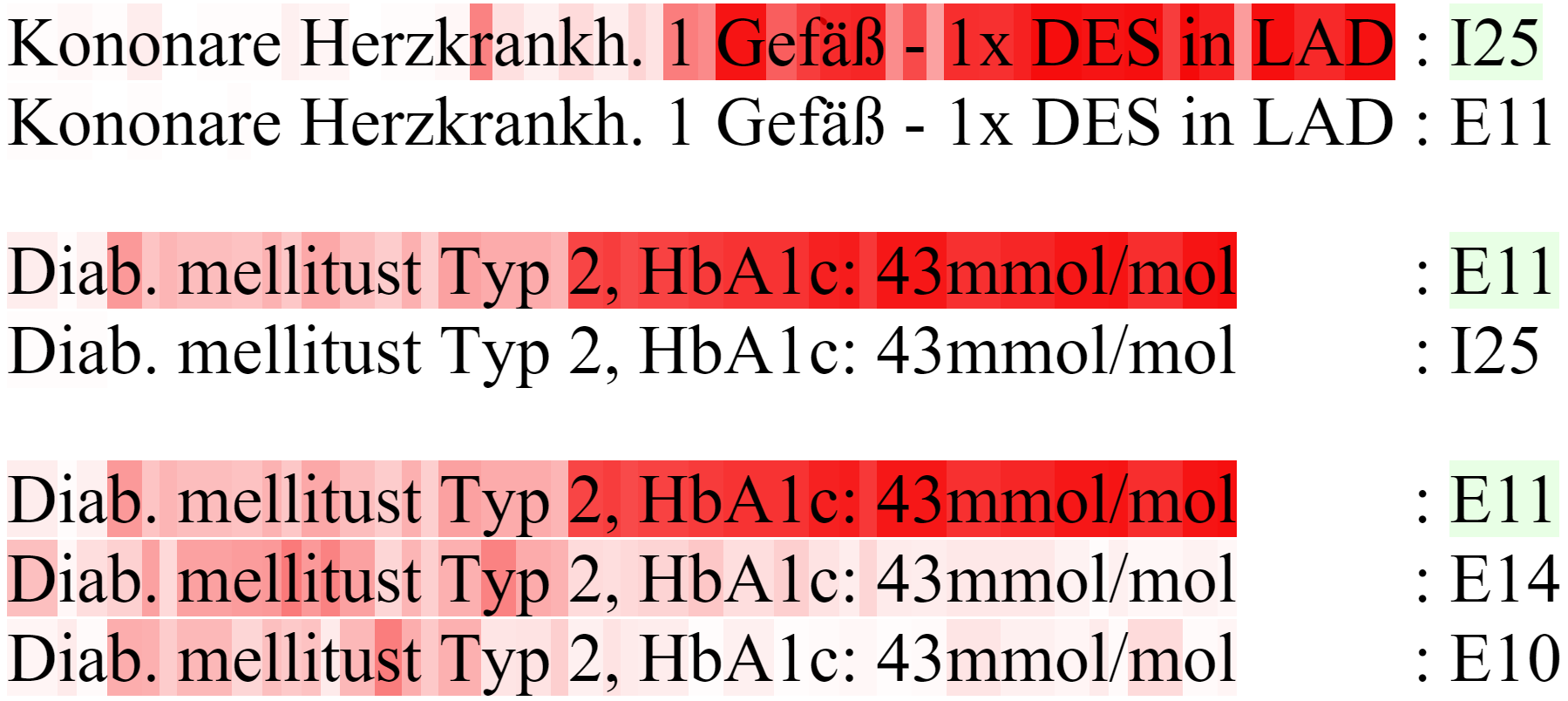}
    \caption{Activation heat map with respect to a given class at a certain character position.}
  	\label{fig:heatMap}
  \end{center}
\end{figure}

What can be seen at the top of Figure~\ref{fig:heatMap} is the network stimulation for the correct class I25 (chronic ischaemic heart disease) for the German input ``Kononare Herzkrankh. 1 Gefäß – 1 x DES in LAD'' (coronary heart disease. 1 vessel – 1 x DES in LAD). The intensity level corresponds to the class level probability for a given code at a given \mbox{position} and varies between zero and one. A low-level intensity therefore corresponds to a low probability for the class shown on the right in the figure. This snippet exhibits the typing mistake ``Kononare'' instead of ``Koronare'' (coronary), an ad-hoc abbreviation ``Herzkrankh.'' for ``Herzkrankheit'' (heart disease) and two acronyms. ``DES'' stands for ``drug-eluting stent'' and ``LAD'' for ``left anterior descending (artery)''. It is principally the last part of this input string that strongly supports the code I25. In contrast, the same string has a very low feedback activation for the ICD-10 class E11 (type 2 diabetes mellitus), as expected.

The centre of Figure~\ref{fig:heatMap} displays the result of the experiment in reverse. We see the network activation for the input ``Diab. mellitust Typ 2, HbA1c: 43 mmol/mol'' (diabetes mellitus type 2, HbA1c: 43 mmol/mol) and its corresponding correct code E11. Again, there are ad-hoc abbreviations ``Diab.'' for ``Diabetes'' and the typing mistake ``mellitust'' (correct ``mellitus''). With the occurrence of the digit ``2'' in the input string, the network responds with a very high activation at this position for type 2 diabetes mellitus. ``HbA1c'' (glycated haemoglobin) is also an important diabetes biomarker. Conversely, there is very little feedback activity for the class I25, resulting in low probability values for the whole input sequence.

At the bottom of Figure~\ref{fig:heatMap}  the input activation for the same string is contrasted with E10 (type 1 diabetes mellitus) and E14 (unspecified diabetes mellitus). As seen before with the appearance of the character ``2'', there is strong evidence for E11. Nevertheless, for E10 and E14, there are clearly observable activation levels. This inspection supports our assumption that, for certain ICD-10 code sections, the network will intermix classes if there is no clear consistent exclusive manual coding. Nevertheless, the network tries to interpret the input as well as possible with respect to the training data.

\section{Conclusion and Outlook}
\label{subsec:conclusionAndOutlook}
We presented an approach for the assignment of short clinical problem list entries to ICD-10 three-digit codes using three different NN architectures: a shallow NN (fastText), a recurrent NN (LSTM) and a transformer-based architecture (RoBERTa) which performed best with an overall macro-averaged F1-score of 0.88. The fastText baseline evaluated with a macro-averaged F1-score of 0.83, the LSTM approach modelling the problem purely on character-level reached a macro-averaged F1-score of 0.84. In a character-level based heat map, we visualized LSTM network activations to a given input. Thus, we could trace classification decisions back. TP, FP and FN analyses revealed that the trained network suffered from coding inconsistencies (e.g., E14 versus E11). 

The preliminary experiments presented in this paper emphasize the potential of secondary use scenarios of annotated semi-structured clinical real-world data for building NN-based applications such as the assignment of disease codes. Taking advantage of such resources is needed for data-driven applications such as the generation of clinical BERT models~\citep{Alsentzer2019a} for languages other than English, like German, where to the best of the authors knowledge a model is not available by now. There is yet an unexploited potential of NNs to bridge the gap between standardized code entries and the language used in the routine of clinical documentation and communication. Identifying and leveraging annotated language resources as presented in this paper could be a step in this direction.

\section*{Acknowledgments}
\label{sec:acknowledgment}
This study was approved by the ethics committee of the Medical University of Graz (30-496 ex 17/18).

\small
\bibliographystyle{unsrtnat}
\bibliography{references}

\end{document}